\documentclass[letterpaper, 10 pt, conference]{ieeeconf}  

\IEEEoverridecommandlockouts                              

\usepackage{graphics} 
\usepackage{epsfig} 
\usepackage{mathptmx} 
\usepackage{times} 
\usepackage{amsmath} 
\usepackage{amssymb}  

\usepackage{threeparttable}
\usepackage{graphicx}
\usepackage{multirow}
\usepackage{bbding}

\usepackage{amssymb}
\usepackage{pifont}
\newcommand{\cmark}{\ding{51}}%
\newcommand{\xmark}{\ding{55}}%

\usepackage[noadjust]{cite}

\title{\LARGE \bf
SocialFormer: Social Interaction Modeling with Edge-enhanced Heterogeneous Graph Transformers for Trajectory Prediction
}

\author{Zixu Wang$^{1, 2*}$, Zhigang Sun$^{3}$, Juergen Luettin$^{1}$, Lavdim Halilaj$^{1}$
\thanks{$^{1}$Zixu Wang, Juergen Luettin, and Lavdim Halilaj are with Robert Bosch GmbH
        {\tt\small \{firstname.lastname\}@bosch.com}}%
\thanks{$^{2}$Zixu Wang is with Technical University of Munich
        {\tt\small \{firstname.lastname\}@tum.de}}%
\thanks{$^{3}$Zhigang Sun is with Bosch Center for Artificial Intelligence
        {\tt\small zhigang.sun3@cn.bosch.com}}%
\thanks{* Work done during an internship at Robert Bosch GmbH}
}

\begin{document}

\maketitle
\thispagestyle{empty}
\pagestyle{empty}

\begin{abstract}
Accurate trajectory prediction is crucial for ensuring safe and efficient autonomous driving. However, most existing methods overlook complex interactions between traffic participants that often govern their future trajectories. 
In this paper, we propose SocialFormer, an agent interaction-aware trajectory prediction method that leverages the semantic relationship between the target vehicle and surrounding vehicles by making use of the road topology. 
We also introduce an edge-enhanced heterogeneous graph transformer (EHGT) as the aggregator in a graph neural network (GNN) to encode the semantic and spatial agent interaction information.
Additionally, we introduce a temporal encoder based on gated recurrent units (GRU) to model the temporal social behavior of agent movements.
Finally, we present an information fusion framework that integrates agent encoding, lane encoding, and agent interaction encoding for a holistic representation of the traffic scene.
We evaluate SocialFormer for the trajectory prediction task on the popular nuScenes benchmark and achieve state-of-the-art performance.

\end{abstract}

\section{INTRODUCTION}

Accurately predicting the future trajectory of nearby vehicles is crucial for ensuring the safety of autonomous driving. 
To achieve this goal, it is important to consider all important factors that govern the future trajectory of a traffic participant. 
These factors include the road map and topology, agent dynamics,
as well as agent interactions~\cite{Huang2022ASO}.
Maps contain important information, including lane boundaries, and possible routes to drive for a given vehicle. Agent dynamics provides information about agents' past movements and direction and may indicate possible future movements. Interactions between agents include the type of maneuver such as lane change, lane keeping, car following, yielding, and giving right-of-way. Furthermore, there exist intricate interactions among agents, where communication plays a role in anticipating the behavior of other agents. Examples include inquiries such as: ``Will the agent behind slow down to let me change the lane?" or ``Will the vehicle stop at the pedestrian crossing to let me cross the lane?" or ``Will the agent stop before the traffic light turns red?". 
In order to model such intricate social agent interactions, it is important to consider the agents movements and understand their relationship~\cite{Ding2023IncorporatingDK,hu2019interaction,sur2022domain,kumar2021interaction,rella2021decoder,jia2023hdgt}. Unfortunately, most methods in trajectory prediction are not able to consider such information.

Several methods~\cite{cui2019multimodal, chai2019multipath, mo2020recog, yuan2021agentformer, gao2020vectornet, liang2020learning, deo2022multimodal} consider road maps and agent dynamics, yet they do not account for the representation of social interactions among agents. 
While certain approaches \cite{li2019grip, jeon2020scale} and \cite{gao2020vectornet, zhao2021tnt, 9700483} adopt homogeneous and heterogeneous graphs, respectively, to model interactions, they often neglect semantic relationships between agents. The manner in which one vehicle influences the trajectory of another is closely tied to their semantic relation within the road network, such as whether a vehicle is ahead of the ego vehicle (\textit{longitudinal}), on an adjacent lane (\textit{lateral}), or on intersecting lanes (\textit{intersecting}). Although recent work~\cite{zipfl2022relation} incorporates such information, it lacks a complete representation and does not consider all relevant information of the map.



\begin{figure}
\centering
\includegraphics[width=\linewidth]
{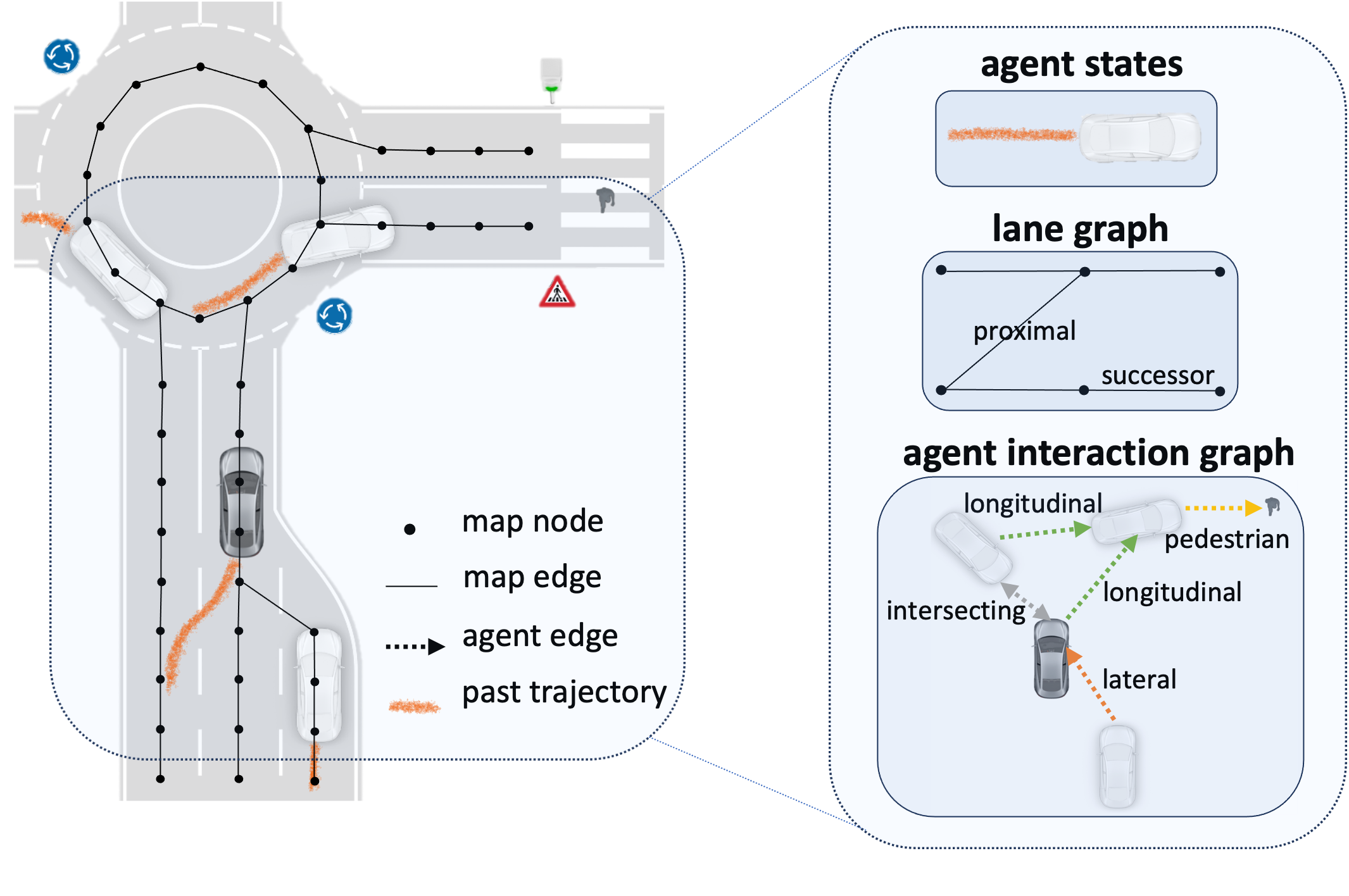}
\vspace{-0.8cm}
\caption{
Schematic illustration of the traffic scene.
The future trajectory of a vehicle depends on many factors, including agent states, road topology with possible driving directions and lane changes, and the influence of nearby vehicles. The latter includes past trajectory, relation type, distance, speed, and right of way, among others.
}
\label{fig: Schematic illustration of heterogeneous driving graph}
\vspace{-0.5cm}

\end{figure}

To tackle these challenges, we propose SocialFormer, a novel approach to trajectory prediction that takes into account social agent interactions, leveraging semantic relationships between agents. To model these interactions, we introduce the edge-enhanced heterogeneous graph transformer (EHGT), which extends the capabilities of the heterogeneous graph transformer (HGT) by addressing its deficiency in incorporating edge attributes. EHGT is capable of encoding both the type and attributes of relationships in heterogeneous graphs, effectively representing the traffic scene and agents. Temporal behavior is modeled using GRUs, and we introduce a fusion module that integrates encodings of the target agent, surrounding agents, their interactions, and lane information. Finally, a multi-layer perceptron (MLP) is employed to predict multiple trajectories for the target vehicle.
In summary, our main contributions are:

\begin{itemize}

\item A novel trajectory prediction architecture that leverages the intricate semantic, spatial, and temporal relationships among agents. This architecture comprises an interaction-aware dynamic heterogeneous graph encoder incorporating interactions between agents with temporal modeling. Additionally, a fusion module is employed to combine various encodings, establishing a comprehensive understanding of the traffic scene for the decoder.

\item A novel edge-enhanced heterogeneous graph transformer (EHGT) operator, capable of encoding edge attributes of heterogeneous graphs.

\item We demonstrate the effectiveness of our approach on the widely-used nuScenes benchmark and conduct several ablation studies to validate our method's performance.





\end{itemize}



\section{RELATED WORK}\label{sec: RELATED WORK}


\subsection{Raster-based Scene Representation}
Raster-based approaches commonly utilize Birds-Eye View (BEV) images as a representation for the map, agents, and related information~\cite{cui2019multimodal, berkemeyer2021feasible}. 
Approaches such as~\cite{bansal2018chauffeurnet,hong2019rules} encode diverse information into a spatial grid, encompassing the history of static and dynamic entities as well as semantic map details. 
MultiPath~\cite{chai2019multipath} employs a deep structure comprising 15 input channels of information, effectively encapsulating both the static and dynamic aspects of the scene. 
However, the rasterization process invariably leads to information loss, including geometric and temporal details. 

\subsection{Graph-based Scene Representation}
Graph-based methods offer the advantage of explicit modeling of scene entities and their interactions within driving scenarios~\cite{halilaj2021knowledge}, potentially leading to improved performance in trajectory prediction.
Approaches like VectorNet~\cite{gao2020vectornet} and its extentions~\cite{zhao2021tnt, gu2021densetnt, huang2022multi, liu2023laformer} use a hierarchical graph to represent the traffic scene. Additionally, these methods incorporate vectorized scene context information extracted from high-definition (HD) maps to enrich their representations.
Compared to VectorNet, the lane graph in LaneGCN~\cite{liang2020learning} and FRM~\cite{park2023leveraging} is only sparsely connected. While graph-based representations provide more explicit modeling, these methods mainly rely on homogeneous graphs, limiting their capacity to represent the semantic content of the traffic scene.



\subsection{Heterogeneous Graph-based Representation}
Several approaches try to model different maps and agents with heterogeneous graphs. 
HEAT~\cite{9700483} extends Graph Attention Networks (GAT)~\cite{velivckovic2017graph} to deal with both heterogeneity and edge features for interaction modeling.  In this method, edge type and edge attributes are concatenated as edge features. Subsequently, this edge feature is concatenated with the source node feature to generate a new node feature used for calculating the attention coefficient and subsequent aggregation. Grimm \textit{et al.}~\cite{grimm2023holistic} use a holistic graph-based method combining time variant information in a single graph, instead of using separate information for each time-step like~\cite{sheng2022graph, cao2021spectral}. This approach is among the first to model the entire encoding step in a single graph for trajectory prediction. 
However, the number of GNN layers must correspond to the need for passing information from the first to the last agent node of the trajectory, which will potentially cause an over-smoothing problem.

\begin{figure*}[ht!]
\centering
\includegraphics[width=\linewidth]{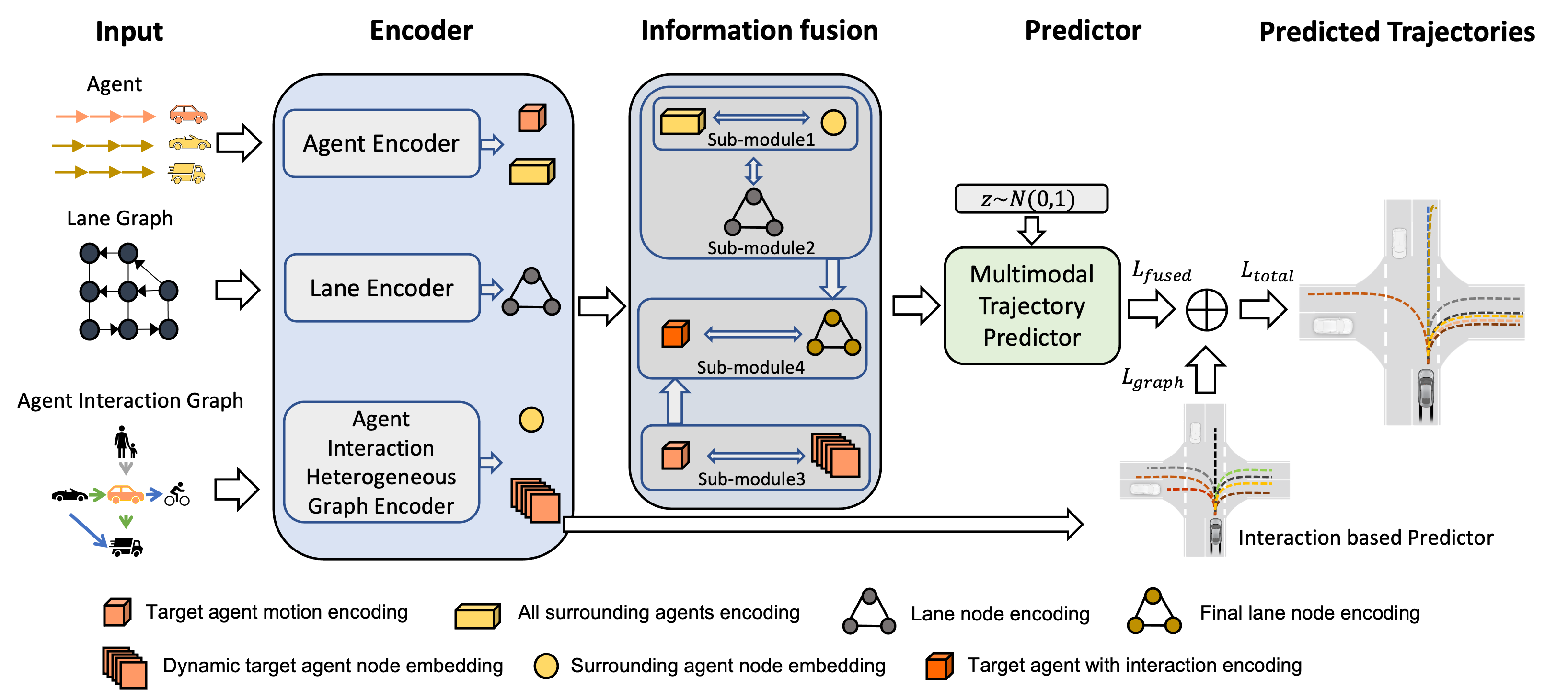}
\caption{Overview of proposed SocialFormer. Agent states, lane graphs, and the interactions between agents in the form of heterogeneous graphs are encoded by the specific encoder. An information fusion module is used to generate the holistic latent representation of the traffic scene. Finally, the predictor outputs possible future trajectories of the target agent.}
\label{fig: Overview of proposed approach}
\end{figure*}

\begin{table}[t]
\centering
\vspace{0.2cm}
\caption{Comparison of different trajectory prediction methods}
\label{table: comparison table}

\resizebox{\linewidth}{!}{
\begin{threeparttable}
\begin{tabular}
{p{2.1cm}||p{0.8cm}|p{0.8cm}|p{0.6cm}|p{0.3cm}|p{0.3cm}||p{0.7cm}|p{0.7cm}|p{0.6cm}}

\hline
\multirow{3}{*}{Method} & \multicolumn{5}{|c||}{Data representation}  &
\multicolumn{3}{c}{Learning approaches}\\
\cline{2-9}

   ~ & \multirow{2}{*}{Map} & \multirow{2}{*}{Agent} & \multicolumn{3}{|c||}{A-A\tnote{0}} & \multirow{2}{*}{Map} & \multirow{2}{*}{Agent} & \multirow{2}{*}{A-A\tnote{0}}\\
   
\cline{4-6}  

  ~ & ~ & ~ & Form & Rel\tnote{1} & EA\tnote{2} & ~ & ~ & ~ \\

\hline 
MTP \cite{cui2019multimodal} &  Image & Image Vector & - & -& -& CNN & CNN & CNN \\ 

Multipath \cite{chai2019multipath}  &  Image  & Image & Image & -& - & CNN &CNN & CNN\\

ReCoG \cite{mo2020recog} & Image  & Vector   & HoE\tnote{4}& - & - & CNN & RNN &GNN\\

AgentFormer \cite{yuan2021agentformer}&  Image &  Vector & - & -& - & CNN & AF\tnote{9}& AF\tnote{9}\\

VectorNet \cite{gao2020vectornet} &  HeN\tnote{5} & HoN\tnote{3}  & - & -& - & MLP & MLP &GNN \\

LaneGCN \cite{liang2020learning}  &  HoN\tnote{3} HeE\tnote{6} &  Vector & - & -& - & GNN & CNN FPN\tnote{7}&ATT\tnote{8}\\

PGP \cite{deo2022multimodal}  &  HoN\tnote{3}  & Vector  & - & - & - & GRU &GRU & -\\

HDGT \cite{jia2023hdgt}  &  HoN\tnote{3} & HeN\tnote{5}  & HeE\tnote{6} & - & \Checkmark & GT\tnote{10} &GT\tnote{10}& GT\tnote{10}\\

GRIP \cite{li2019grip}&  - & HoN\tnote{3} & HoE\tnote{4}& - & - &GCN & GCN\\

HEAT \cite{9700483}  &  Image & HeN\tnote{5} & HeE\tnote{6} & - & \Checkmark& CNN &HEAT & HEAT\\

Relation \cite{zipfl2022relation}  & NA  & HoN\tnote{3} & HeE\tnote{6} &  \Checkmark &  \Checkmark & NA & GNN & GNN\\

LAformer \cite{liu2023laformer}  & Vector  & Vector  & - & -& -& ATT\tnote{8} &ATT\tnote{8} & -\\
 
FRM \cite{park2023leveraging}& HoN\tnote{3} HeE\tnote{6}  & Vector & - & -& -& GNN & MLP & FRM\tnote{11}\\
 
HoliGraph \cite{grimm2023holistic}  &  HoN\tnote{3} HeE\tnote{6} & HoN\tnote{3} & HeE\tnote{6} & \Checkmark & \Checkmark& GNN &GNN &GNN \\


SocialFormer(Ours) &  HoN\tnote{3} HeE\tnote{6}& HeE\tnote{6} Vector & HeE\tnote{6} &\Checkmark &\Checkmark & GNN &GNN &EHGT \\
 
\hline 

\end{tabular}

$^0$A-A: Agent-Agent, 
$^1$Rel: Relation type,
$^2$EA:~Edge Attribute,
$^3$HoN:~Homogeneous Node,
$^4$HoE: Homogeneous Edge,
$^5$HeN: Heterogeneous Node,
$^6$HeE:~Heterogeneous Edge,
$^7$FPN:~Feature Pyramid Network,
$^8$ATT:~Attention,
$^9$AF:~AgentFormer,
$^{10}$GT:~Graph Transformer,
$^{11}$FRM:~Future Relationship Module,
$^{12}$-: Not modelled, \Checkmark: Modelled

\end{threeparttable}
}
\end{table}

\subsection{Agent Interaction Modeling}
Several methods have been proposed for modeling the interaction between agents. 
TNT~\cite{zhao2021tnt} utilizes a hierarchical heterogeneous graph to represent interactions. Initially, each object is represented by a sub-graph, which is then combined into a fully connected graph. However, learning interactions from such a fully connected graph can be challenging. AgentFormer~\cite{yuan2021agentformer} models interactions between agents in time and interactive dimensions with a transformer mechanism. However, this method also does not directly represent relationships between agents. GRIP~\cite{li2019grip} explicitly represents interactions between agents using a graph-based approach, where agents close to each other are connected by edges. However, it does not consider edge features such as distance and speed difference.
SCALE-Net~\cite{jeon2020scale} employs an edge-featured homogeneous graph to represent agent interactions, encoding the relative states between connected agents. 
HEAT~\cite{9700483} represents interactions as an edge-featured heterogeneous graph but still ignores the semantic relationships among the agents. In \cite{grimm2023holistic}, Grimm \textit{et al.} uses the edge type \textit{attent} to represent the interaction between agents. In~\cite{zipfl2022relation, grimm2023heterogeneous}, semantic information regarding agent interaction is represented by relation types \textit{longitudinal}, \textit{lateral}, and \textit{intersecting}, ignoring relation with pedestrian. However, these approaches still do not comprehensively represent social interactions.




As shown in Table \ref{table: comparison table}, which offers a detailed comparison of different trajectory prediction methods with a particular focus on social interaction modeling between agents, many methods either ignore agent social interaction or only use limited information.
This omission hinders capturing the insight from the complex traffic scenarios between traffic participants. In this paper, we proposed  SocialFormer, a novel approach with a focus on agent social interaction with heterogeneous graphs to address these shortcomings. Incorporating this information from complex traffic scenarios, our architecture is able to learn and then predict trajectories considering the interactive intention of traffic participants.

\section{METHODOLOGY}\label{sec: METHODOLOGY}
We first provide an overview of our approach as shown in Fig.~\ref{fig: Overview of proposed approach}. 
Initially, we encode all agent information and lane graphs into latent embeddings. Subsequently, we employ a dynamic heterogeneous graph encoder to encode the agent interaction graph. The information fusion module then integrates the diverse encodings with four sub-modules, forming a holistic representation. Finally, the decoder utilizes the comprehensive representation to predict multi-modal trajectories. 


\subsection{Lane Graph Representation}
We adopt a graph representation for the map in line with \cite{deo2022multimodal}. Center lines of lanes are represented as a directed graph, denoted as \textit{G = \{V, E\}}. 
Each node $v$ denotes a sequence of pose vectors
 \begin{equation}
 f_{1:N} ^{v} = [f_{1}^{v}, f_{2}^{v}, ...,f_{N}^{v}],
 \end{equation}
 where every pose is characterized by:
 \begin{equation}
f_{n} ^{v} = [x_{n}^{v}, y_{n}^{v},\theta_{n}^{v}, \sideset{}{_n^v}{\mathop{flag}}_{stopline}, \sideset{}{_n^v}{\mathop{flag}}_{crosswalk}],
 \end{equation}
 where $x_{n}^{v}$, $y_{n}^{v}$ are the local coordinates for the $n$-th pose and $\theta_{n}^{v}$ is its yaw angle. $flag$ indicates whether the pose lies on a stop line or crosswalk.

Regarding node inter-relationships, there are two edge types: \textit{successors} and \textit{proximal}. The \textit{successors} edge ensures continuity to the next node along a lane, maintaining a legitimate trajectory. In contrast, the \textit{proximal} edge represents legal lane changes between neighboring lanes traveling in the same direction (see Fig.~\ref{fig: Schematic illustration of heterogeneous driving graph}).

\subsection{Agent Representation}\label{sec: agent representation}
Agents, categorized as either human or vehicle, are represented as sequences of temporal vectors:
\begin{equation}
f_{n}^{T} = [f_{n}^{t = -4}, f_{n}^{t = -3}, ..., f_{n}^{t = 0}],
\end{equation}
where $n$ specifies the individual agent. 
The index $t$ spans from the earliest observable frame to the present scene (2-second historical states with a 2 Hz sampling rate). Each temporal vector $f_{n}^{t}$ represents state features as 
 \begin{equation}
 f_{n}^{t} = [x^{t}, y^{t}, vel^{t}, acc^{t}, yaw rate^{t}],
 \end{equation}
 indicating the location, velocity, acceleration, and yaw rate of $n$-th agent at $t$.

\subsection{Lane and Agent Encoder}\label{sec: Lane and Agent Encoder in approach}
In order to model the temporal dynamics of the agents, we first use an MLP followed by GRUs to encode lane node features $f_{1:N} ^{v}$, target agent's states $f_{target}^{T}$ and surrounding agent's states $f_{surr}^{T}$. The output from each GRU serves as the embedding for the lane node ($h_{lane}$), the target agent ($h_{target}$), and surrounding agents ($h_{surr}$).

\subsection{Agent Interaction Representation}
When driving, humans rely on a complex interplay of factors to prioritize which surrounding vehicles or agents are most relevant. Based on the work of \cite{zipfl2022towards} and \cite{mlodzian2023nuscenes}, we identify the inter-agent relationships by four distinct edge types: \textit{Longitudinal}, \textit{Intersecting}, \textit{Lateral},  and \textit{Pedestrian} with three edge attributes $f_{edge}^{(s,t)}  = [Distance, PathDistance, EdgeProbability]$ 
(see Fig.~\ref{fig: Schematic illustration of heterogeneous driving graph}), 
where $Distance$ quantifies the Euclidean distance,  $PathDistance$ specifies the distance traversed along a specific path, and $EdgeProbability$ provides the likelihood of a particular relation. Notably, the \textit{Pedestrian} edge type does not possess the $PathDistance$ attribute.


\subsection{Agent Interaction Dynamic Encoding}

\begin{figure}[t]
\centering
\includegraphics[width=\linewidth]
{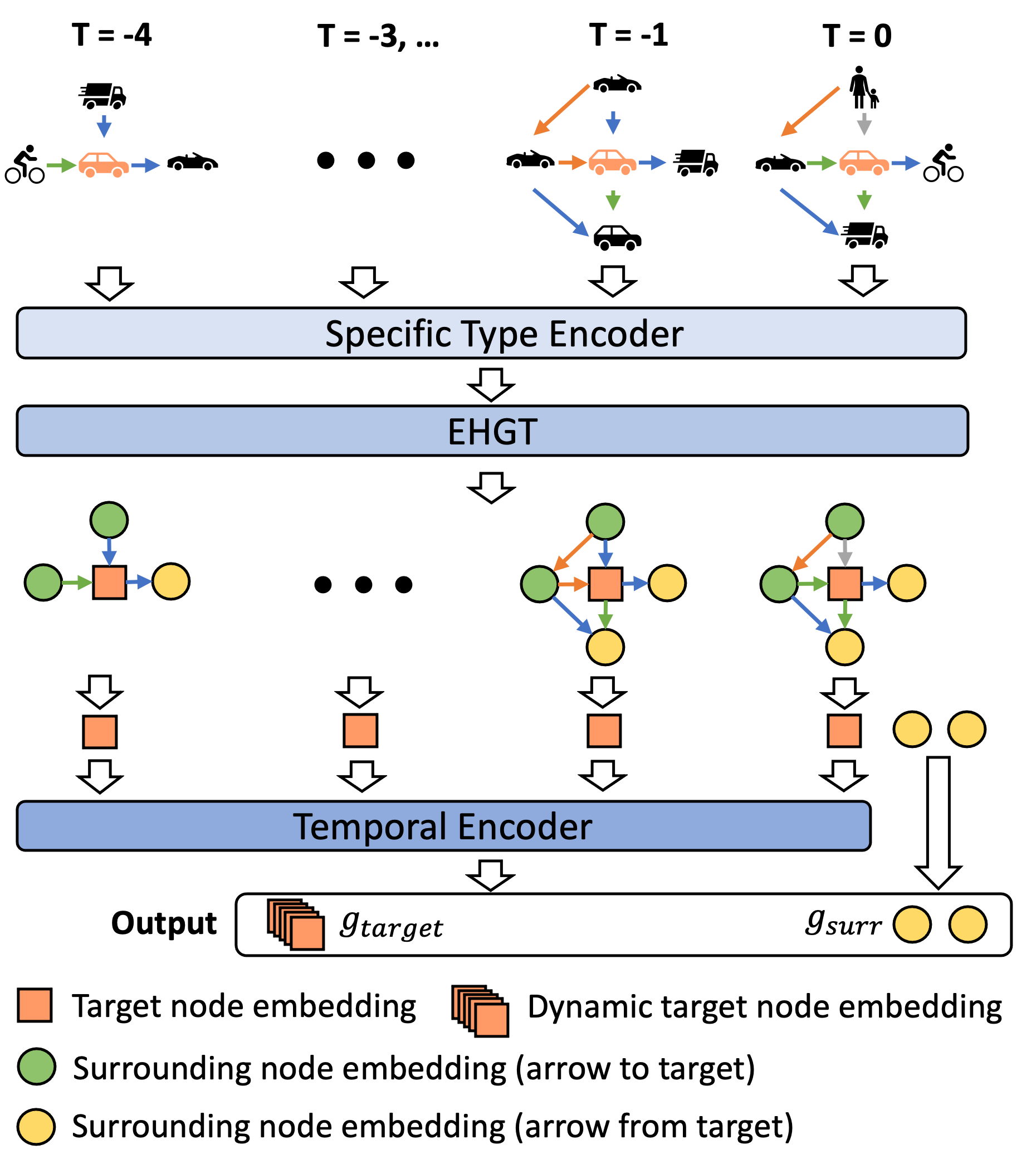}
\caption{Illustration of Agent Interaction Dynamic Heterogeneous Graph Encoder: It consists of three parts: specific type encoder, EHGT, and temporal encoder.}
\label{fig: Dynamic Heterogeneous Knowledge Graph Encoder}
\vspace{-0.3cm}
\end{figure}

To model the different dynamic behavioral patterns of different agent types (e.g., vehicle or pedestrian), we use distinct MLPs as specific type encoder to encode them, as shown in 
Fig. \ref{fig: Dynamic Heterogeneous Knowledge Graph Encoder}.
Subsequently, we leverage an EHGT
to encode 2-second historical and current agent interaction graphs.
Finally, a temporal encoder is deployed to capture temporal changes in the agent interactions.


While Heterogeneous Graph Transformer (HGT)~\cite{hu2020heterogeneous} has proven effective in modeling relational data within heterogeneous graphs, its inability to incorporate edge attributes constrains its application in scenarios where the crucial information conveyed by attributes associated with the edges is required. To tackle this problem, we propose EHGT, an extension of HGT, which is able to incorporate edge attributes into the mutual attention and message-passing stages, offering a more comprehensive graph representation.
In mutual attention, target node \textit{t} is projected into a Query vector, while the source node \textit{s} is mapped into a Key vector:
\begin{align} \label{equ: Key and Query in HGT}
Q^{i}_{(t)} &= Q\_Linear^{i}_{\tau(t)}(H^{l-1}[t]) \\
K^{i}_{(s)} &= K\_Linear^{i}_{\tau(s)}(H^{l-1}[s]),
\end{align}
where $H^{(l-1)}$ is the input from the previous layer and $i$ refers to the specific attention head. Each node type has a unique linear projection. When computing the attention head, we employ an edge attribute matrix $W_{\phi (e)}^{Eg\_Attr}$ :
\begin{equation} \label{equ: Edge HGT method2: ATT}
\small
ATT\_head^i(s,e,t) = (  K^i(s)W_{\phi (e)}^{ATT} W_{\phi (e)}^{Eg\_Attr} Q^{i}(t)^{T})\cdot \frac{\mu _{<\tau (s), \phi (e), \tau (t) >}  }{\sqrt{d} },  
\end{equation}
where $W_{\phi (e)}^{ATT}$ is a matrix specific to the edge type, designed to capture the semantic relations $\phi (e)$ between nodes. Furthermore, $\mu$ stands as an adaptive parameter, signifying the importance of each relation triple $\left \langle source\ node, edge, target\ node\right \rangle$, while $d$ represents the dimensionality of the vector. Then $h$ attention heads are concatenated, and the $Softmax$ function is applied to determine the final attention weights for each relation triple:
\begin{equation} \label{equ: attention weights in HGT}
ATT_{EHGT}(s,e,t) = \mathop{Softmax}\limits_{s\in N(t)}\left(\parallel_{i\in \left [ 1,h \right]} ATT\_head^i(s,e,t)  \right ).
\end{equation}

The message-passing runs in parallel with the mutual attention computation. To compute \textit{i}-th message head $MSG\_head^{i}(s,e,t)$, the representation of the source node from the preceding layer $H^{l-1}[s]$ is projected by a node type encoder $M\_Linear$. 
Subsequently, it is multiplied by the edge attribute matrix $W_{\phi (e)}^{Eg\_Attr}$ and matrix $W^{MSG}_{\phi (e)}$  to integrate the edge dependency:
\begin{equation} \label{equ: Edge HGT method2: MSG}
 MSG\_{head}^{i}(s,e,t) = M\_Linear^{i}_{\tau(s) }(H^{l-1}[s])W_{\phi (e)}^{Eg\_Attr} W^{MSG}_{\phi (e)}.
\end{equation}
We then concatenate $h$ message nodes for each node pair to obtain $MSG_{EHGT}(s,e,t)$:
\begin{equation} \label{equ: message in HGT}
    MSG_{EHGT}(s,e,t) = \parallel _{i\in[1,h] } MSG\_{head}^{i}(s,e,t).
\end{equation}

In the subsequent aggregation phase, messages from all source nodes \textit{s} are aggregated to the target node \textit{t}. The attention vector in Eq.~\ref{equ: attention weights in HGT} serves as the weight to average the corresponding messages from source nodes in Eq.~\ref{equ: message in HGT}. 
This results in the updated $\tilde{H}^{(l)}[t]$:
\begin{equation}
\tilde{H}^{(l)}[t] = \oplus_{\forall s \in N(t) }(ATT_{EHGT}(s,e,t)\cdot MSG_{EHGT}(s,e,t)).
\end{equation}
After this, $\tilde{H}^{(l)}[t]$ contains comprehensive information from the target node \textit{t}'s neighbors and their associated relations. The target node \textit{t} is then remapped to its type-specific distribution and augmented with a residual connection from the previous layer:
\begin{equation}
H^{(l)}[t] = A\_Linear_{\tau(t )} (\sigma (\tilde{H}^{(l)}[t] ))+ H^{(l-1)}[t].
\end{equation}

EHGT enables the processing of heterogeneous graphs with edge attributes. We employ EHGT to process agent interaction heterogeneous graphs. In practical scenarios, it is typically only the direct neighbor of a target vehicle that significantly influences its behavior. Therefore, we restrict our aggregation for surrounding agents within one hop. 




To capture the evolution of traffic scenes over time, we employ a temporal encoder with GRU, which encodes the target node embeddings sourced from all observed traffic scenes in temporal order, resulting in a dynamic target node graph embedding $g_{target}$. Simultaneously, we preserve the embedding of surrounding nodes $g_{surr}$ from the last observed traffic scene (see Fig. \ref{fig: Dynamic Heterogeneous Knowledge Graph Encoder}), specifically, those nodes that emanate arrows from the target agent, to ensure that we do not lose sight of potential agents affecting prediction.

\subsection{Information Fusion}
As described above, several encodings are utilized to comprehensively represent the dynamic and interactive information within traffic scenes:

\begin{itemize}
\item Target Agent Motion Encoding ($h_{target}$): This captures the movements of the target agent.

\item Surrounding Agents Encoding ($h_{surr}$): This encoding provides information about all non-target agents within a traffic scene.

\item Lane Node Encoding ($h_{lane}$): This gives insights into the structured lanes in the scene.

\item Dynamic Target Node Graph Embedding ($g_{target}$): This embedding represents the target agent's interactions and relations within a dynamic graph.

\item Surrounding Agent Interactive Embedding ($g_{surr}$): 
This captures how the surrounding agents interact with each other and with the target agent.
\end{itemize}

\begin{figure}
\centering
\vspace{0.2cm}
\includegraphics[width=\linewidth]
{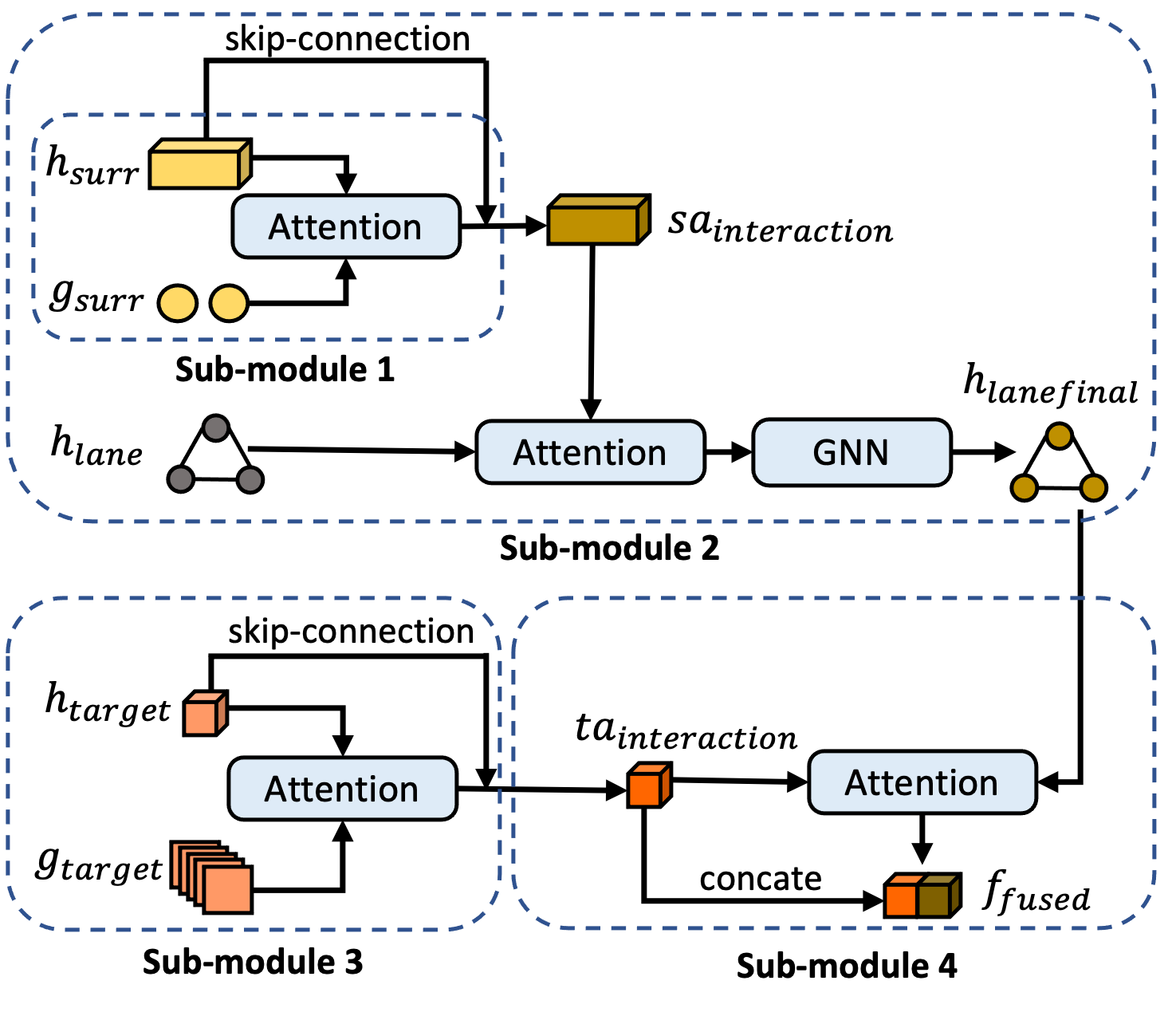}
\caption{Information Fusion Module: Four sub-modules combine different encodings to form a comprehensive encoding for the predictor. The output is $f_{fused}$.
}
\label{fig: Information fusion all}
\end{figure}

To optimally integrate the insights from these encodings, we employ an information fusion module with four sub-modules, as shown in Fig. \ref{fig: Information fusion all}.

\subsubsection{Sub-module 1: Surrounding Agents and their Interactions}

To extract meaningful representations that highlight the most influential surrounding agents and understand the significance of their interactions, we deploy a cross-attention mechanism to weigh and understand which relationships of the surrounding agents are most relevant in the given context. 
The output of the Sub-module~1 $sa_{interaction}$ denotes the surrounding agents with their encoded interactions:
\begin{equation}\label{equ:sub-module1}
    sa_{interaction} = CrossAttention(h_{surr}, g_{surr}) + h_{surr}.
\end{equation}

Besides, we add a skip-connection to ensure the transfer of the original information across the cross-attention layers and to stabilize the model, especially when the interaction graph is sparse.


\subsubsection{Sub-module 2: Surrounding Agents with Interaction and Lane Information}

Incorporating a cross-attention mechanism between $sa_{interaction}$ and $h_{lane}$ ensures that the lane node embeddings are refined based on the presence and behavior of agents close to them. We then use GNN to update lane node embedding with other connected lane nodes. After this, we have the final encoding $h_{lane final}$ for lane information, which is both context-aware (in terms of agent interactive behaviors) and structurally informed (in terms of lane configurations): 
\begin{equation}\label{equ:sub-module2}
    h_{lanefinal} = GNN(CrossAttention(h_{lane}, sa_{interaction})).
\end{equation}


    

\subsubsection{Sub-module 3: Target Agent and its dynamic Interaction}

Similar to Sub-module 1, we like the target agent to be aware of the broader context in which it operates, especially considering its interactions with nearby agents. 
We therefore refine  $h_{target}$ by fusing it with $g_{target}$:
\begin{equation}\label{equ:sub-module3}
    ta_{interaction} = CrossAttention(h_{target}, g_{target}) + h_{target},
\end{equation}
where $ta_{interaction}$ indicates the target agent with interaction encoding that is richly informed by its inherent attributes and social interactions with other agents.

\subsubsection{Sub-module 4: Target Agent with Interaction Encoding and Lane Final Encoding}

We apply the attention mechanism to weigh the relevance of various features from $h_{lane final}$ to $ta_{interaction}$:
\begin{equation}\label{equ:sub-module4}
\small
    f_{fused} = Concate(ta_{interaction},CrossAtt(ta_{interaction}, h_{lanefinal})).
\end{equation}

Intuitively, this allows our module to focus on the most pertinent lane-related information while considering a target agent's interaction. 
Then, we concatenate $ta_{interaction}$ with the attention output to ensure that it not only captures the weighted information from the lanes but also preserves the raw interaction details of the target agents. Finally, we obtain the holistic fused encoding $f_{fused}$, which is the final encoding of the information fusion module.

This step is crucial as it is the final information fusion process. It provides a complete encoding that includes the specific details from the target agent's view and its interactions with others and the lanes around it.

\subsection{Multimodal Trajectory Predictor}
The fused information is then processed by the multi-modal trajectory predictor. We concatenate the Gaussian distribution variable $z$ with the fused encoding $f_{fused}$, so that the decoder is equipped to generate diverse motion profiles, accounting for the inherent uncertainty. 
Then we use a MLP to output $k$ mode future trajectories $\hat{Y}_{1:t_{f}}^{k}$:
\begin{equation}
\hat{Y}_{1:t_{f}}^{k} = MLP(concate(f_{fused}, z_k)).
\end{equation}
We then adopt K-means clustering and output the cluster centers as the final output of $K$ predictions
$[\hat{Y}_{1:t_{f}}^{1}, \hat{Y}_{1:t_{f}}^{2}, ..., \hat{Y}_{1:t_{f}}^{k}]$. 
Besides, we also decode the dynamic target node graph embedding $g_{target}$ directly to trajectories $\tilde{Y}_{1:t_{f}}^{k}$. 
Based on these two outputs, we train our decoder with winner-takes-all average displacement error. In this way, we have two losses: fused regression loss $\mathcal{L}_{fr}$ and graph regression loss $\mathcal{L}_{gr}$ and a combined loss of them:
\begin{equation} \label{equ: combined loss}
    \mathcal{L} = \lambda_1 \mathcal{L}_{fr} + \lambda_2 \mathcal{L}_{gr}.
\end{equation}
with
\begin{equation} \label{equ: fused regression loss}
    \mathcal{L}_{fr} = min_{k}\frac{1}{t_f} \sum_{t = 1}^{t_f} \left \|\hat{Y}_{t}^{k} - Y_{t}^{gt}\right\|_{2},
\end{equation}
\begin{equation} \label{equ: graph regression loss}
   \mathcal{L}_{gr} = min_{k}\frac{1}{t_f} \sum_{t = 1}^{t_f} \left \|\tilde{Y}_{t}^{k} - Y_{t}^{gt}\right\|_{2}.
\end{equation}
The combined loss $\mathcal{L}$ brings together the strengths of both individual losses, pushing the model to learn from the rich spatial, semantic, and temporal features embedded in $\mathcal{L}_{fr}$ and $\mathcal{L}_{gr}$, ensuring a comprehensive
learning process.

\section{EXPERIMENT}\label{sec: EXPERIMENT}
We evaluate our proposed SocialFormer for the nuScenes prediction challenge, which is to predict the 6-second future trajectory of the target vehicle, given 2-second historical data with a sampling rate of 2Hz. 
A performance comparison with state-of-the-art approaches and adequate ablation studies are demonstrated.



\subsection{Dataset}
We use the widely-used public nuScenes dataset~\cite{caesar2020nuscenes} and the traffic scene graph dataset nSTP~\cite{mlodzian2023nuscenes}, which is a heterogeneous graph dataset representing the nuScenes dataset. It models all scene participants, road elements, and their semantic and spatial relationships. 
In total, 32186 training samples and 9041 validation samples are utilized for our experiments.

\subsection{Metrics}
We use Average displacement error ($\text{ADE}_\text{k}$) and miss rate ($\text{MR}_\text{k}$) as  evaluation metrics. $\text{ADE}_\text{k}$ is to quantify average L2 distance errors between the predicted trajectory and the ground truth (GT), whereas $\text{MR}_\text{k}$ denotes the percentage of cases for the maximum L2 distance error is larger than 2 meters over $k$ most likely predictions.


\subsection{Experimental Setup}
We implemented and trained our SocialFormer method in the PyTorch framework~\cite{paszke2019pytorch} using the Nvidia V100 GPU server. All graphs are handled with PyTorch Geometric (PyG)~\cite{Fey/Lenssen/2019}, a popular library for developing GNNs for structured data applications. We apply the AdamW optimizer~\cite{loshchilov2017decoupled} with an initial learning rate of 0.001. 


\section{RESULT}\label{sec: RESULT}

\begin{table}[t]
\vspace{0.15cm}
    \centering
    \caption{
    Results on nuScenes benchmark for different models
    }
    \label{table: Comparison to other approaches on nuScenes}
    \resizebox{\linewidth}{!}{
        \begin{threeparttable}
            \begin{tabular}{c||*{3}{c|}c}
                \hline
                Model & ADE$_{5}$ & MR$_{5}$ & ADE$_{10}$ & MR$_{10}$ \\
                \hline 
                Multipath \cite{chai2019multipath} & 2.32 & - & 1.96 & -\\
                
                CoverNet \cite{phan2020covernet} & 1.96 & 0.67 &1.48 & -\\
                
                P2T \cite{deo2020trajectory}& 1.45 & 0.64 & 1.16 & 0.46\\
                
                Trajectron++ \cite{salzmann2020trajectron++} & 1.88 & 0.70 & 1.51 & 0.57\\ 
                
                AgentFormer \cite{yuan2021agentformer} & 1.86 & - & 1.45 & -\\
                Autobot \cite{girgis2021latent} & 1.37 & 0.62 & 1.03  & 0.44\\ 
                
                SG-Net \cite{wang2022stepwise}& 1.86 & 0.67 & 1.40 & 0.52\\ 
                
                PGP \cite{deo2022multimodal} & \textbf{1.30} & \underline{0.61} & \underline{1.00} & \textbf{0.37}\\ 

                HeteGraph \cite{grimm2023heterogeneous} & - & - & 1.29 & 0.57\\
                
                SocialFormer (Ours) & \underline{1.32} & \textbf{0.58} & \textbf{0.98} & \underline{0.39}\\ 
                \hline 
            \end{tabular}
        \end{threeparttable}
    }
\end{table}

A comparison of the results with SocialFormer and state-of-the-art methods is shown in Table \ref{table: Comparison to other approaches on nuScenes}, which
presents the results obtained on the nuScenes validation set. Metrics are the minimum average displacement error ADE and miss rate MR for the top 5 and top 10 predictions, respectively. We highlight the best results in boldface and the second-best results in underlined.
Our method achieves the best prediction accuracy for the 
 $\text{MR}_\text{5}$  and $\text{ADE}_\text{10}$ metrics. 
For the metrics $\text{ADE}_\text{5}$ and $\text{MR}_\text{10}$, our SocialFormer  outperforms other state-of-the-art models except PGP~\cite{deo2022multimodal}.
This demonstrates the effectiveness of our SocialFormer in the trajectory prediction task. 

It is worth noting that not all traffic scenes exhibit all four defined relation types. Moreover, there are 959 scenes (i.e., 2.33\% of all samples) that lack any defined semantic relations, which is because the interactions between two agents do not fit the predefined relation categories, other agents are located more than one lane away from the target agent, or even there are no or very sparse agents surrounding to the target agent. However, our SocialFormer still shows robust performance in predicting the future trajectory of the target agent.

\subsubsection{Qualitative Results}

\begin{figure}
\centering
\vspace{0.15cm}
\includegraphics[width=\linewidth]
{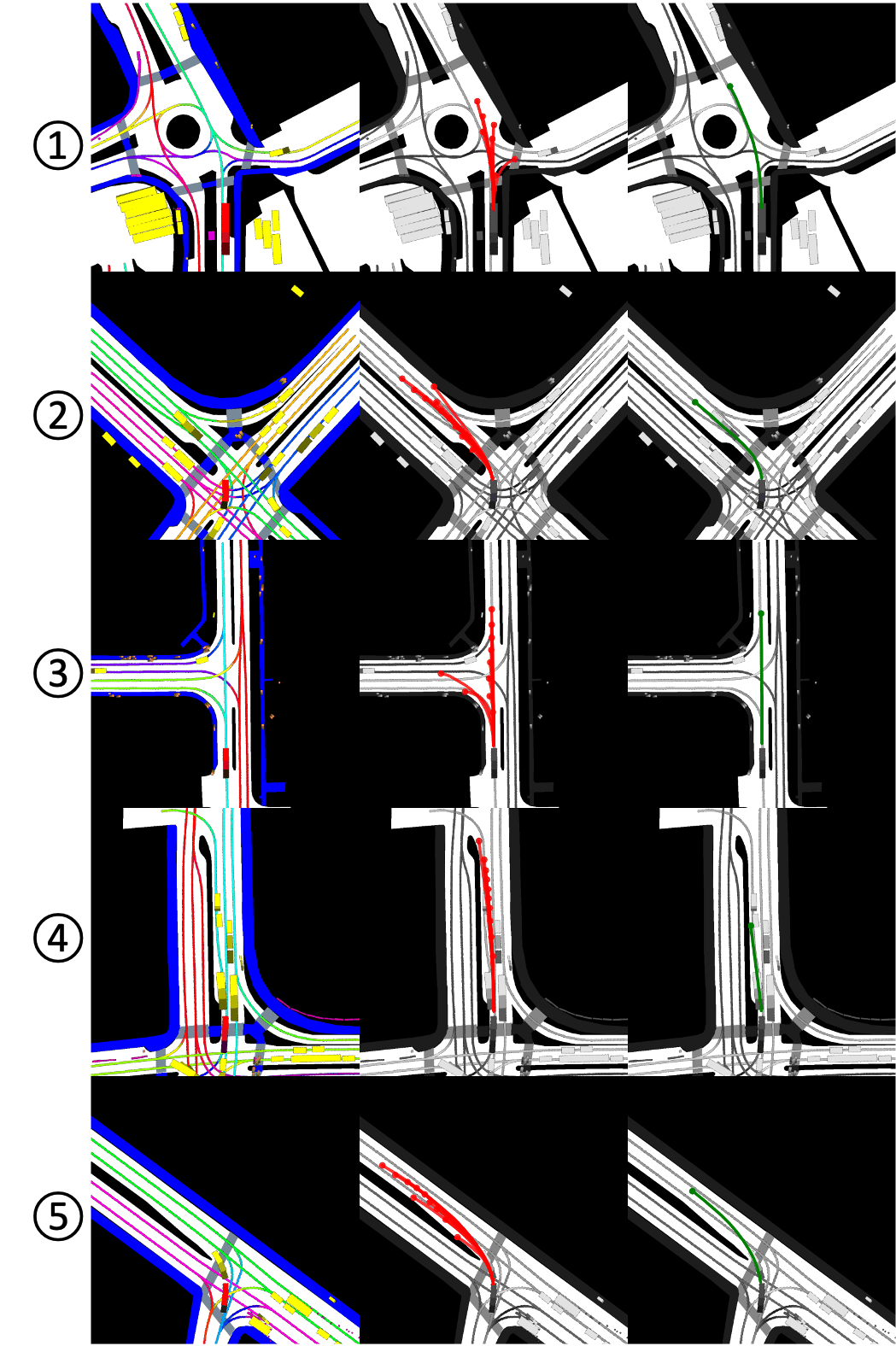}
\caption{Illustration of the qualitative result in various traffic scenarios. Left column: HD maps and tracks. Middle column: Top 10 most likely predictions. Right column: Ground truth.}
\label{fig: Qualitative result}
\end{figure}

In Fig. \ref{fig: Qualitative result}, we present qualitative results of our model in different scenarios, with each row representing a unique scenario. 
In scenario \textcircled{4}, our model accurately generates predictions in straightforward driving with longitudinal diversity. Scenario \textcircled{1} illustrates a complex roundabout scene, where our model predicts diverse routes, with most predictions being correct. In scenarios \textcircled{3}, accurate predictions are generated at intersections. Meanwhile, in scenarios \textcircled{2} and \textcircled{5}, our model accurately predicts various routes in left-turn situations with a possibility of lane change.

\subsubsection{Ablation Study}
We conduct an ablation study to validate the effect of each component in our architecture,
including different widely-used GNN operators, temporal encoder, and diverse information fusion strategies. 
The baseline model predicts trajectories solely based on map and agent data, excluding interactions between agents. In Table \ref{table: Ablation study on edge attribute and GNN operators}, we compare the performance of both homogeneous  (i.e., GAT and GCN) and heterogeneous GNN operators (i.e., heterogeneous graph attention network (HAN) and RGAT). We also explore $\text{HGT}_{\text{concate}}$, which simply concatenates edge attributes (EA) to node features.

We observe an improvement on all metrics with the integration of edge attributes using our EHGT, underscoring the significance of effectively modeling edge attributes. 

\begin{table}[htbp]
    \centering
    \vspace{0.2cm}
    \caption{Ablation study on edge attribute and GNN operators}
    \label{table: Ablation study on edge attribute and GNN operators}
    \resizebox{\linewidth}{!}{
        \begin{threeparttable}
            \begin{tabular}{*{7}{c|}c}
                \hline
                EA & GNN & ADE$_5$ & MR$_5$& FDE$_5$ & ADE$_{10}$ & MR$_{10}$ & FDE$_{10}$ \\
                \hline 

                \xmark & HGT & 1.35 & \textbf{0.58} & 2.72 & 0.99 & 0.40 & 1.70 \\ 

                \xmark & GAT & 1.35 & 0.59 & 2.71 & 0.99 & 0.40 & 1.71 \\ 

                \xmark & GCN & 1.36 & 0.60 & 2.70 & 0.99 & 0.41 & 1.71 \\ 

                \xmark & HAN & 1.35 & 0.60 & 2.67 & \textbf{0.98} & 0.40 & 1.68 \\ 

                \cmark & RGAT & 1.36 & 0.60 & 2.68 & 0.99 & \textbf{0.39} & 1.68 \\

                \cmark & HGT$_{concate}$  & 1.35 & \textbf{0.58} & 2.67 & 0.99 & \textbf{0.39} & 1.69 \\
                
                \cmark & EHGT(Ours) & \textbf{1.32} & \textbf{0.58} & \textbf{2.64} & \textbf{0.98} & \textbf{0.39} & \textbf{1.67} \\ 

                \hline 
            \end{tabular}
        \end{threeparttable}
    }
\end{table}

Table \ref{table: Ablation study on fusion methods and temporal encoder} demonstrates a performance improvement with the inclusion of a temporal encoder, which indicates the significance of social interactions in the temporal dimension.
However, the skip connections do not yield a substantial benefit. This observation may be attributed to the relatively shallow depth of our model.

\begin{table}[htbp]
    \centering
    \vspace{0.18cm}
    \caption{Ablation study on fusion methods and temporal encoder}
    \label{table: Ablation study on fusion methods and temporal encoder}
    \resizebox{\linewidth}{!}{
        \begin{threeparttable}
            \begin{tabular}{*{7}{c|}c}
                \hline
                
                Temp. Enc. & Fusion & ADE$_5$ & MR$_5$& FDE$_5$ & ADE$_{10}$ & MR$_{10}$ & FDE$_{10}$ \\
                \hline 

                \xmark & Att. & 1.35 & \textbf{0.57} & 2.73 & 0.99 & 0.40 & \textbf{1.64} \\ 

                \xmark & Att.,Skip & 1.35 & 0.58 & 2.72 & 0.99 & 0.40 & \textbf{1.64} \\ 

                \cmark & Att. & 1.33 & 0.58 & \textbf{2.64} & 0.99 & 0.40 & 1.67 \\
                
                \cmark & Att.,Skip & \textbf{1.32} & 0.58 & \textbf{2.64} & \textbf{0.98} & \textbf{0.39} & 1.67 \\ 

                \hline 
            \end{tabular}
        \end{threeparttable}
    }
\end{table}

\subsubsection{Sensitivity Analysis of the Hyper-Parameters}


We conducted an empirical study by varying the weights of the fused regression loss in Eq.~\ref{equ: combined loss}. The results in Table~\ref{table: Loss weight} reveal that the combination of $\lambda_1$ = 1 and $\lambda_2$ = 0.5 yields the best performance across all metrics. 
These findings underscore the efficacy of the encoding from the dynamic heterogeneous graph encoder in enhancing the model's understanding of agent interaction graphs. 
Acting as part of the auxiliary task, the graph loss ensures that the loss is effectively propagated into the graph encoder within the intricate architecture. 

\begin{table}[htbp]
    \centering
    \caption{Comparison of different loss weight combination}
    \label{table: Loss weight}
        \begin{threeparttable}
            \begin{tabular}{c|*{6}{c|}*{1}{c}}
                \hline
                
                $\lambda_{1}$ & $\lambda_{2}$ & ADE$_5$ & MR$_5$ & FDE$_5$ & ADE$_{10}$ & MR$_{10}$ & FDE$_{10}$ \\
                 
                \hline 
                
                1 & 0 & 1.34 & 0.60 & 2.65 & 0.98 & 0.39 & 1.68 \\

                1 & 0.2 & 1.34 & 0.59 & 2.65 & 0.99 & 0.41 & 1.67 \\
                
                \textbf{1} & \textbf{0.5} & \textbf{1.32} & \textbf{0.58} & \textbf{2.64} & 0.98 & 0.39 & 1.67 \\

                1 & 1 & 1.36 & 0.61 & 2.66 & 0.99 & 0.41 & 1.69 \\
                \hline 
            \end{tabular}
        \end{threeparttable}
\end{table}

\section{CONCLUSION}\label{sec: CONCLUSION}

This paper introduced SocialFormer, a novel trajectory prediction architecture that integrates agent interaction with heterogeneous graphs. 
We proposed the edge-enhanced graph transformer (EHGT) to encode the semantic relations and their attributes and a GRU-based temporal encoder to extract semantic relations in the temporal domain. A fusion module was proposed to integrate the diverse agent, lane, and interaction encodings in a more comprehensive manner.
Experiments demonstrate that our model achieves state-of-the-art performance on the popular nuScenes dataset, even when sparse or no semantic information exists in some traffic scenes.
It also indicates that social interaction modeling could improve other trajectory prediction methods.
Future work will consider employing the information on traffic signs and lights and incorporating anchor paths to remove unrealistic predictions and to avoid mode collapse.

\bibliographystyle{IEEEtran}
\bibliography{Bibliography}

\end{document}